\documentclass[]{article}

\PassOptionsToPackage{numbers,sort&compress}{natbib}
\usepackage[preprint]{workshop2026}

\usepackage[T1]{fontenc}
\usepackage{microtype}
\usepackage[utf8]{inputenc} 

\usepackage{amsmath,amsthm,amsfonts,amssymb}

\DeclareMathAlphabet{\mathmybb}{U}{bbold}{m}{n}

\usepackage{hyperref}
\usepackage[capitalise]{cleveref}
\usepackage{enumitem}
\usepackage{url}

\usepackage{xcolor}
\usepackage{subcaption}
\usepackage{booktabs}

\usepackage{tikz}
\usepackage{pgfplots}
\usepgfplotslibrary{groupplots}
\pgfplotsset{compat=1.18}

\usepackage{algorithm}
\usepackage{algpseudocode}

\usepackage{mdframed}

\newcommand{\ba}{\boldsymbol{a}}

\newcommand{\bc}{\boldsymbol{c}}

\newcommand{\bo}{\boldsymbol{o}}

\newcommand{\bx}{\boldsymbol{x}}

\newcommand{\bz}{\boldsymbol{z}}

\newcommand{\bJ}{\boldsymbol{J}}

\makeatletter
\newsavebox{\@brx}
\newcommand{\llangle}[1][]{%
  \savebox{\@brx}{\(\m@th{#1\langle}\)}%
  \mathopen{\copy\@brx\kern-0.5\wd\@brx\usebox{\@brx}}}
\newcommand{\rrangle}[1][]{%
  \savebox{\@brx}{\(\m@th{#1\rangle}\)}%
  \mathclose{\copy\@brx\kern-0.5\wd\@brx\usebox{\@brx}}}
\makeatother

\crefname{conjecture}{Conjecture}{Conjectures}
\crefname{claim}{Claim}{Claims}

\theoremstyle{definition}

\crefname{assumption}{Assumption}{Assumptions}
\crefname{appendix}{Appendix}{Appendices}

\title{Behavioral Monitoring of JEPA World Models \\ with Jacobian Centroids}
\author{
  Thomas Walker\thanks{Correspondence: \texttt{thomas.walker@rice.edu}} \\
  Rice University\\
  \And
  Randall Balestriero \\
  Brown University \\
  \And
  Richard Baraniuk \\
  Rice University
}

\begin{document}

\maketitle

\begin{abstract}
Detecting failures in World Model (WM)-based planning requires monitoring whether the model is behaviorally aligned with the current task, which in turn requires studying its internal representations. 
Here, we show that centroids---sub-component Jacobian row-sums---effectively identify the behavioral properties of WMs, complementing traditional activation-based knowledge signals. 
The centroids of a model are easily computed through Jacobian vector products and characterize how the model organizes the geometry of its input space, yielding an efficient perspective on internal representations, including the generation of task-relevant saliency maps. Evaluated on continuous control tasks using JEPA WMs, this behavioral view reveals a structural dissociation, where the encoder correctly represents the goal while the predictor remains behaviorally unresponsive. 
This failure mode directly predicts planning failure before any action is taken, allowing for goal resampling to recapture out-of-distribution success. 
Moreover, centroid-based methods outperform baseline methods as distribution-shift detectors. 
Together, these tools yield a behavioral monitoring stack that is operational and consequential under distribution shifts.
\end{abstract}

\section{Introduction}

World Models (WMs) trained from observational data are a promising route to sample-efficient planning~\cite{lecun2022path,maes2026leworldmodel}, but deploying them reliably requires monitoring their internal representations at runtime.
Otherwise, plans suffering from representational misalignment could lead to practical failures.
The Linear Representation Hypothesis (LRH)~\cite{park2023linear,elhage2022superposition} assumes that features are encoded linearly in the activations of the intermediate layers of a model, making cosine similarity and linear probing natural alignment measures.
Indeed, prior WM monitoring approaches have used reconstruction error~\cite{foundation2025bimanual} and conformal uncertainty envelopes over activations~\cite{seo2025unisafe,angelopoulos2022conformal} for avoiding out-of-distribution failures.

In this paper, we consider centroids as a complementary signal.
Under the spline theory of deep learning~\cite{balestrieroSplineTheoryDeep2018}, centroids parameterize the input space geometry of models~\cite{balestrieroGeometryDeepNetworks2019}.
For a sub-component of a model, per-sample centroids are computed as the row sum of the corresponding input-output Jacobian~\cite{balestrieroGeometryDeepNetworks2019}.
The Linear Centroids Hypothesis (LCH)~\cite{walker2026linear} establishes that centroids share the same linear structure as activations while more directly capturing behavioral properties.

We show that the centroids of Joint Embedding Predictive Architecture (JEPA) WMs~\cite{lecun2022path} reveal their behavioral properties, allowing for the construction of saliency maps (\Cref{fig:saliency_maps}), out-of-distribution detectors (\Cref{tab:auc_summary}), and adaptive planning protocols (\Cref{tab:gate_resample}).
For example, planning failure can be preempted by identifying when the encoder correctly represents the goal, but the predictor remains behaviorally unresponsive to it (\Cref{tab:quadrants}).






\section{The LeWorldModel and its Behavioral Structure}\label{sec:features_wms}

We focus on the LeWorldModel~\cite{maes2026leworldmodel} instantiation of JEPA WMs~\cite{lecun2022path}, a two-stage architecture comprising a Vision Transformer (ViT)~\cite{dosovitskiy2021an} encoder $e:\mathbb{R}^d\to\mathbb{R}^h$ and an action-conditioned autoregressive predictor $p:\mathbb{R}^h\times\mathbb{R}^m\to\mathbb{R}^h$, each built from stacked transformer layers~\cite{vaswaniAttentionAllYou2017} with multi-layer perceptron (MLP) blocks.
Given an observation $\bo_t$, the encoder produces an embedding $\bz_t = e(\bo_t)$, and the predictor forms an embedding $\widehat{\bz}_{t+1} = p(\bz_t, \ba_t)$.
The WM is trained offline to minimize $\|\widehat{\bz}_{t+1} - \bz_{t+1}\|_2^2$ (with SIGReg to prevent collapse~\cite{maes2026leworldmodel}).
At planning time, given the goal $\bo_\text{goal}$, a CEM planner~\cite{rubinstein2004cross} optimizes an $H$-step action sequence to minimize $\|\bz_\text{goal} - \widehat{\bz}_H\|_2$ (only $K$ steps of the plan are executed before re-planning).

In the following, we compare centroid (behavioral) and activation (knowledge) structures in WMs.
For a given input, centroids are computed layerwise as the row-sum of the input-output Jacobians of the MLP blocks~\cite{balestrieroGeometryDeepNetworks2019}, and activations are extracted layerwise from the outputs of the MLP blocks.
Concretely, for a given sub-component of a WM $g$ and a given input $\bx$ to that sub-component, the centroid is given by $\bc=\left(\bJ_g[\bx]\right)^\top\mathbf{1}$ and the intermediate activation is given by $g(\bx)$; where $\bJ_g[\bx]$ denotes the input-output Jacobian of $g$ at $\bx$.
Within the spline theory of deep learning~\cite{balestrieroSplineTheoryDeep2018}, the centroid $\bc$ parametrizes the region of the input space partition induced by $g$ that $\bx$ occupies~\cite{balestrieroGeometryDeepNetworks2019}.

We answer three questions about the behavioral structure of the WMs' encoder and predictor, focusing on Push-T~\cite{chi2023diffusionpolicy,chi2024diffusionpolicy} (with replications in TwoRoom~\cite{sobal2026learning} and Reacher~\cite{tassa2018deepmind} in \Cref{fig:tworooms,fig:reacher}, respectively).

\paragraph{Do the predictor's representations track the encoder's?}

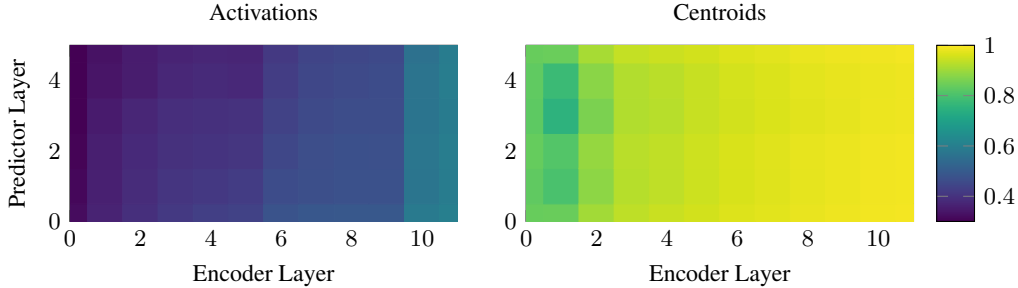
\begin{figure}[ht]
    \centering
    \begin{tikzpicture}
    \begin{axis}[
        name=heatmap_act,
        width=0.48\textwidth, height=0.28\textwidth,
        enlargelimits=false,
        tick label style={font=\small}, label style={font=\small}, title style={font=\small},
        xlabel={Encoder Layer}, ylabel={Predictor Layer}, title={Activations},
        point meta min=0.3, point meta max=1, colormap/viridis,
        xmin=0, xmax=11, ymin=0, ymax=5,
    ]
    \addplot[matrix plot*, mesh/cols=6, point meta=explicit]
        table[x=enc_layer, y=pred_layer, meta=act_r2, col sep=comma]{data/exp1_heatmaps.csv};
    \end{axis}
    \begin{axis}[
        name=heatmap_cent,
        at={(heatmap_act.outer east)}, anchor=outer west, xshift=0.5cm,
        width=0.48\textwidth, height=0.28\textwidth,
        enlargelimits=false,
        tick label style={font=\small}, label style={font=\small}, title style={font=\small},
        xlabel={Encoder Layer}, title={Centroids},
        point meta min=0.3, point meta max=1, colorbar, colormap/viridis,
        xmin=0, xmax=11, ymin=0, ymax=5
    ]
    \addplot[matrix plot*, mesh/cols=6, point meta=explicit]
        table[x=enc_layer, y=pred_layer, meta=cent_r2, col sep=comma]{data/exp1_heatmaps.csv};
    \end{axis}
    \end{tikzpicture}
    \caption{
        \textbf{A WM's predictor behaves in alignment with its encoder.} 
        Each predictor layer's behavioral sensitivity (centroids, right) is linearly predictable from the encoder's. 
        Activations (left) between the encoder and predictor are less linearly predictable.
    }
    \label{fig:r2_heatmaps}
\end{figure}

Using ridge regression between layer-wise principal component analysis projections of centroids and activations from expert trajectories, we find that each predictor layer's behavioral sensitivity is linearly predictable (as measured by $R^2$ values) from the encoder's (see \Cref{fig:r2_heatmaps}).
For both centroids and activations, the alignment increases with encoder depth and remains approximately constant with predictor depth.
The alignment is stronger in the case of centroids.
This illustrates that behavioral alignment is a structural property of the jointly trained WM.

\paragraph{Do expert actions trace smooth paths through representation space?}

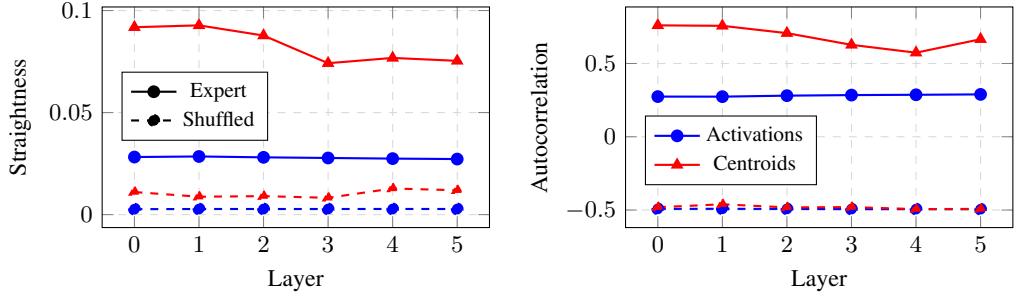
\begin{figure}[ht]
    \centering
    \begin{tikzpicture}
        \begin{groupplot}[
            group style={
                group size=2 by 1,
                horizontal sep=1.8cm,
            },
            width=0.48\textwidth,
            height=4.5cm,
            xtick={0,1,2,3,4,5},
            xlabel={Layer},
            tick label style={font=\small},
            label style={font=\small},
            scaled y ticks=false,
            grid=major,
            grid style={dashed, gray!30},
            y tick label style={
                /pgf/number format/.cd,
                fixed,
                precision=3,
                /tikz/.cd,
            },
        ]
        \nextgroupplot[
            ylabel={Straightness},
            legend style={
                at={(0.05, 0.55)},
                anchor=west,
                font=\footnotesize,
            }
        ]
            \addplot[color=blue, mark=*, thick, forget plot]
                table[x=layer, y=act_straightness_expert, col sep=comma]
                {data/exp5_predictor.csv};
            \addplot[color=blue, mark=*, thick, dashed, forget plot]
                table[x=layer, y=act_straightness_shuffled, col sep=comma]
                {data/exp5_predictor.csv};
            \addplot[color=red, mark=triangle*, thick, forget plot]
                table[x=layer, y=cent_straightness_expert_mean, col sep=comma]
                {data/exp5_predictor.csv};
            \addplot[color=red, mark=triangle*, thick, dashed, forget plot]
                table[x=layer, y=cent_straightness_shuffled_mean, col sep=comma]
                {data/exp5_predictor.csv};
    
            \addlegendimage{color=black, mark=*, thick}
            \addlegendentry{Expert}
            \addlegendimage{color=black, mark=*, dashed, thick}
            \addlegendentry{Shuffled}

        \nextgroupplot[
            ylabel={Autocorrelation},
            legend style={
                at={(0.05, 0.35)},
                anchor=west,
                font=\footnotesize,
            }
        ]
            \addplot[color=blue, mark=*, thick, forget plot]
                table[x=layer, y=act_autocorr_expert, col sep=comma]
                {data/exp5_predictor.csv};
            \addplot[color=red, mark=triangle*, thick, forget plot]
                table[x=layer, y=cent_autocorr_expert_mean, y error=cent_autocorr_expert_std, col sep=comma]
                {data/exp5_predictor.csv};
            \addplot[color=blue, mark=*, thick, dashed, forget plot]
                table[x=layer, y=act_autocorr_shuffled, col sep=comma]
                {data/exp5_predictor.csv};
            \addplot[color=red, mark=triangle*, thick, dashed, forget plot]
                table[x=layer, y=cent_autocorr_shuffled_mean, y error=cent_autocorr_expert_std, col sep=comma]
                {data/exp5_predictor.csv};
                
            \addlegendimage{color=blue, mark=*, thick}
            \addlegendentry{Activations}
            \addlegendimage{color=red, mark=triangle*, thick}
            \addlegendentry{Centroids}
            
        \end{groupplot}
    \end{tikzpicture}
    \caption{\textbf{Expert actions trace smooth paths through a WM predictor's feature space.} Straightness and lag-1 autocorrelation both exceed frame-shuffled controls at every predictor layer, for activations and centroids. Centroids are consistently straighter (higher straightness, left) and smoother (higher autocorrelation, right).
    }
    \label{fig:trajectory_metrics}
\end{figure}

We measure two trajectory properties at each predictor layer across five Push-T expert episodes.
\emph{Straightness} is the ratio of end-to-end displacement to total path length, $\|\bc_T-\bc_0\|\,/\,\sum_t\|\Delta\bc_t\|\in[0,1]$, with 1 indicating a perfectly straight path.
\emph{Autocorrelation} is the lag-1 vector autocorrelation of consecutive displacement vectors, $\sum_t (\Delta\bc_t\cdot\Delta\bc_{t+1})/\sum_t\|\Delta\bc_t\|^2$, measuring directional persistence between steps.
The \emph{shuffled} control randomly permutes frames within each episode, preserving the marginal distribution of representations while destroying temporal structure.
Expert trajectories are straighter and more autocorrelated than shuffled controls at every predictor layer, for both centroids and activations (see \Cref{fig:trajectory_metrics}), with centroids consistently smoother than activations.

\paragraph{Is the encoder's behavioral attention task-salient?}

Centroids exist within the input space of the sub-component from which they are computed.
Meaning, centroids computed from sub-components whose input space is the observation space (e.g., the map from the observation space to the output of an intermediate layer) can be directly visualized as saliency maps.
To assess whether the encoder's behavioral attention is task-meaningful and to situate centroids relative to established methods, we compare centroids against Integrated Gradients (IG)~\cite{sundararajan2017ig} and GradCAM~\cite{selvaraju2019gradcam}.
Because IG and GradCAM require a nominated scalar output, we construct $s(\boldsymbol{x}) = \sum_d [\boldsymbol{W}_\text{proj}\,\boldsymbol{z}_\text{CLS}(\boldsymbol{x})]_d$
(where $\boldsymbol{x}_\text{CLS}$ is the final-layer CLS token and $\boldsymbol{W}_\text{proj}$ is the projector MLP) as the simplest natural scalar derived from the encoder's output.
The \emph{centroid} is the input gradient $\nabla_{\boldsymbol{x}} s(\boldsymbol{x})$, obtained in a single forward--backward pass.
IG~\cite{sundararajan2017ig} integrates along a path from a black baseline, requiring multiple forward--backward passes leading to higher computational cost.
GradCAM~\cite{selvaraju2019gradcam} produces a coarse patch-resolution map bilinearly upsampled to the input resolution.

\newcommand{\salcell}[1]{\includegraphics[width=0.23\textwidth]{figures/#1}}
\begin{figure}[ht]
\centering
\setlength{\tabcolsep}{2pt}
\renewcommand{\arraystretch}{0.5}
\begin{tabular}{cccc}
  \small Input &
  \small Centroid &
  \small Integrated Gradients &
  \small GradCAM \\[2pt]
  \salcell{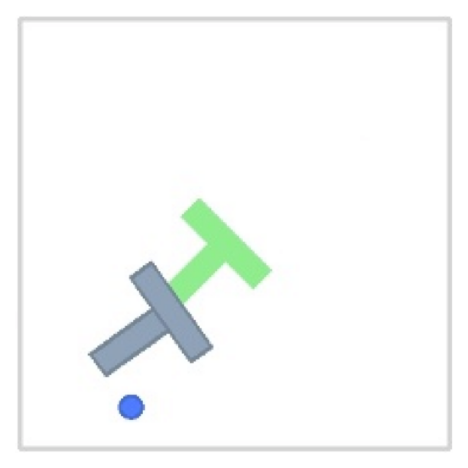} &
  \salcell{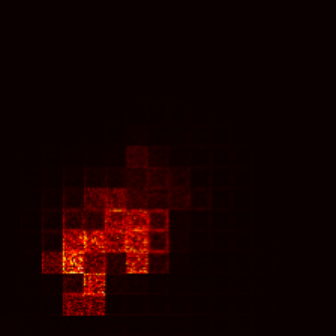} &
  \salcell{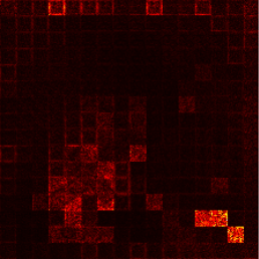} &
  \salcell{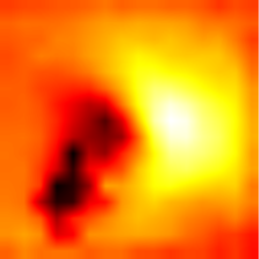} \\
  \multicolumn{4}{c}{\small Frame 68} \\[4pt]
  \salcell{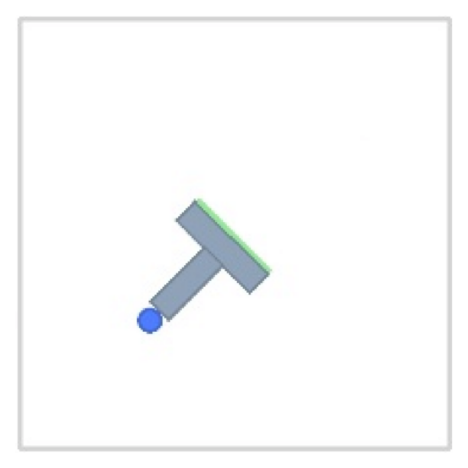} &
  \salcell{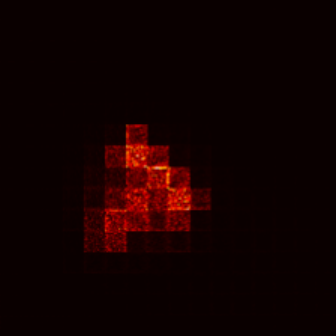} &
  \salcell{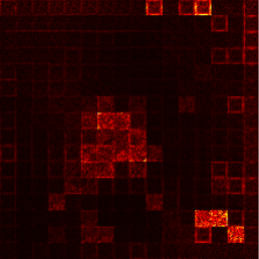} &
  \salcell{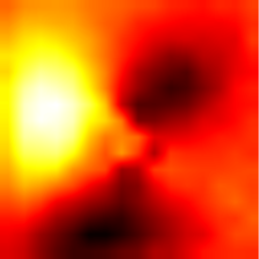} \\
  \multicolumn{4}{c}{\small Frame 108} \\
\end{tabular}
\caption{\textbf{Encoder behavioral attention is efficiently captured using centroids as saliency maps.}
Score target $s(\boldsymbol{x}) = \sum_d [\boldsymbol{W}_\text{proj}\,\boldsymbol{z}_\text{CLS}(\boldsymbol{x})]_d$.
All maps min-max normalised to $[0,1]$ per frame (\texttt{hot} colormap: dark$\to$low, bright$\to$high).
IG~\cite{sundararajan2017ig} ($N{=}100$, black baseline) produce coherent spatial maps.
GradCAM~\cite{selvaraju2019gradcam} (ViT layer~10, no ReLU) is visibly coarser.
Frame~68 captures the approach phase; frame~108 the contact phase.
}
\label{fig:saliency_maps}
\end{figure}

From \Cref{fig:saliency_maps}, all methods concentrate attention on the ball and T-piece, with salience intensifying around the contact region in the later frame---confirming the encoder's behavioral attention is task-specific across methods.
Centroids and IG produce spatially coherent maps; GradCAM is coarser, consistent with its patch-level resolution.
Critically, the centroid achieves this with a single backward pass and no nominated output dimension.

\section{Centroid Signals for Failure Prediction and Adaptive Replanning}\label{sec:failure_prediction}

The behavioral alignment between predictor and encoder (\Cref{fig:r2_heatmaps}) and the smoothness of expert trajectories through behavioral space (\Cref{fig:trajectory_metrics}) motivate using centroid-based signals as runtime monitors.
Intuitively, when the predictor's behavioral sensitivity fails to track the encoder's, we would expect planning failure to be almost imminent.
We explore this directly, comparing behavioral (centroid-based) and knowledge (activation-based) signals for predicting episode outcomes and guiding adaptive replanning.

\paragraph{Experimental setup.}

We evaluate on Push-T and TwoRoom using the trained LeWorldModel with a CEM planner (300 samples, 30 steps, receding-horizon $K=2$), and label each episode as a success or failure based on the environment's feedback.
The \emph{in-distribution} regime sets the goal offset at 25 steps.
Across 5 seeds and 100 episodes per seed, the model achieves a success rate of $88.0 \pm 3.3\%$ on Push-T and $84.4 \pm 3.9\%$ on TwoRoom.
The \emph{out-of-distribution} regime sets the goal offset to 75 steps.
In this regime, the model achieves a success rate of $18.8 \pm 6.8\%$ on Push-T and $34.8 \pm 6.9\%$ on TwoRoom.

We probe six signal families (full definitions in \Cref{sec:signal_defs}): post-execution calibration $c_\text{cen}$ (cosine similarity between the predictor's centroid evaluated on the realized vs.\ predicted next embedding, computed per predictor MLP layer); its activation-space analogue $c_\text{act}$; goal-code cosine $c_\text{goal}$ computed from a centroid dictionary via Orthogonal Matching Pursuit~\cite{pati1993orthogonal} (OMP; omp-cen) and its activation counterpart (omp-act); a sparse autoencoder (SAE)-coded~\cite{bricken2023monosemanticity,templeton2024scaling} centroid dictionary variant (sae-cen); the standard embedding cost $c_\text{emb}$; and a reconstruction-error baseline $c_\text{recon}$ (centroid and activation variants)---the OMP sparse reconstruction error on the training dictionary, implementing the same principle as variational autoencoder (VAE)-based WM monitoring~\cite{foundation2025bimanual} without a trained decoder.
We report AUC for predicting episode success or failure.

\subsection{Failure Detection and Attribution}

\paragraph{Knowledge and behavioral signals converge in-distribution.}

In-distribution, $c_\text{cen}$ and $c_\text{act}$ are closely matched on Push-T ($0.863$ vs $0.876$, $\Delta = -0.013$), with activations marginally ahead; the gap inverts under distribution shift ($c_\text{cen}$ $0.914$ vs $c_\text{act}$ $0.910$, $\Delta = +0.004$).
On TwoRoom the centroid advantage is large in both regimes ($\Delta = +0.154$ in-distribution, $\Delta = +0.096$ out-of-distribution).

\begin{table}[ht]
\centering
\small
\caption{%
\textbf{Failure-prediction AUC across signal families.}
In both regimes, five seeds with 100 episodes each are considered so that mean plus standard deviations are reported.
Per-column best is in \textbf{bold}.
The goal-code cosine spans centroid vs.\ activation and OMP vs.\ TopK SAE~\cite{gaoScalingEvaluatingSparse2025} with $256$ features and $k\!=\!4$.
For $c_\text{cen}$, layer-0 on Push-T and layer-5 on TwoRoom are consistently near-optimal (Push-T gap $<0.02$ AUC in-distribution, $<0.01$ out-of-distribution), providing a practical task-adaptive rule with no post-hoc selection.
}\label{tab:auc_summary}
\begin{tabular}{lcccc}
\toprule
Signal & \multicolumn{2}{c}{In-distribution} & \multicolumn{2}{c}{Out-of-distribution} \\
\cmidrule(lr){2-3}\cmidrule(lr){4-5}
& Push-T & TwoRoom & Push-T & TwoRoom \\
\midrule
\multicolumn{5}{l}{\textit{Reconstruction baseline (VAE-style analogue, \Cref{sec:signal_defs})}} \\[2pt]
$c_\text{recon}$ (centroid)   & $0.550 \pm 0.195$ & $0.708 \pm 0.125$ & $0.623 \pm 0.060$ & $0.791 \pm 0.066$ \\
$c_\text{recon}$ (activation) & $0.592 \pm 0.170$ & $0.700 \pm 0.121$ & $0.632 \pm 0.029$ & $0.776 \pm 0.066$ \\
\midrule
\multicolumn{5}{l}{\textit{Knowledge signals (activation-based)}} \\[2pt]
$c_\text{emb}$ (embedding cost)  & $0.782 \pm 0.096$ & $\mathbf{0.971 \pm 0.026}$ & $0.947 \pm 0.025$ & $0.945 \pm 0.026$ \\
$c_\text{goal}$ (omp-act)       & $0.646 \pm 0.072$ & $0.871 \pm 0.033$ & $0.958 \pm 0.008$ & $0.946 \pm 0.023$ \\
$c_\text{goal}$ (sae-act)  & $0.710 \pm 0.101$ & $0.911 \pm 0.038$ & $0.948 \pm 0.017$ & $0.973 \pm 0.022$ \\
$c_\text{act}$ (emb.\ calib.)    & $\mathbf{0.876 \pm 0.024}$ & $0.756 \pm 0.051$ & $0.910 \pm 0.053$ & $0.817 \pm 0.048$ \\
\midrule
\multicolumn{5}{l}{\textit{Behavioral signals (centroid-based)}} \\[2pt]
$c_\text{cen}$      & $0.863 \pm 0.040$ & $0.910 \pm 0.036$ & $0.914 \pm 0.034$ & $0.913 \pm 0.029$ \\
$c_\text{goal}$ (omp-cen)       & $0.656 \pm 0.073$ & $0.863 \pm 0.036$ & $\mathbf{0.960 \pm 0.008}$ & $0.964 \pm 0.020$ \\
$c_\text{goal}$ (sae-cen)  & $0.681 \pm 0.104$ & $0.923 \pm 0.029$ & $0.951 \pm 0.013$ & $\mathbf{0.978 \pm 0.015}$ \\
\bottomrule
\end{tabular}
\end{table}

\paragraph{Reconstruction-error baselines underperform calibration and goal-code signals.}

Reconstruction baselines $c_\text{recon}$ achieve AUC $0.55\pm0.20$ (centroid) and $0.59\pm0.17$ (activation) in-distribution on Push-T, and $0.71\pm0.12$ / $0.70\pm0.12$ on TwoRoom, which is below all calibration and goal-code signals (\Cref{tab:auc_summary}).
The gap is most pronounced in-distribution (${\sim}0.31$ AUC behind $c_\text{cen}$ on Push-T) and narrows modestly out-of-distribution.
The high variance of $c_\text{recon}$ in-distribution on Push-T also indicates unreliable detection when the model is operating near its training regime.

\paragraph{OMP and SAE are two effective feature-extraction strategies.}

Since the LRH~\cite{park2023linear,elhage2022superposition} and LCH~\cite{walker2026linear} establish that features are encoded linearly in both activations and centroids, sparse feature extraction over these representations is well-motivated.
We compare OMP and SAE as two strategies for building goal-code dictionaries.
Both improve over raw embeddings in-distribution, with gains that differ by task.
The centroid gains are task-concentrated, whereas activation gains are more symmetric.

OMP centroid and OMP activation are essentially tied in-distribution.
With SAE coding, activations modestly outperform centroids on Push-T in-distribution, while centroids maintain a consistent advantage on TwoRoom.
On TwoRoom, SAE centroid dictionaries (sae-cen) are the strongest single goal-code signal; on Push-T, OMP centroid (omp-cen) is the more stable choice (std $0.073$ vs.\ $0.101$) and retains a slight out-of-distribution edge over SAE activation ($0.960$ vs.\ $0.948$).
Out-of-distribution, omp-cen ($0.960 \pm 0.008$) and omp-act ($0.958 \pm 0.008$) are statistically indistinguishable on Push-T ($\Delta = +0.002$).
The behavioral advantage is marginally reliable on TwoRoom (omp-cen $0.964 \pm 0.020$ vs omp-act $0.946 \pm 0.023$, $\Delta = +0.018$).

\paragraph{Layer-wise calibration signal profile is task-dependent.}

$c_\text{cen}$ is computed per predictor MLP layer, yielding the per-layer out-of-distribution AUC profiles shown in \Cref{fig:layer_auc}.
The optimal layer differs markedly by task: on Push-T, the signal peaks at shallow layers (layer-0--layer-1, AUC~$0.928$--$0.929$) and degrades with depth.
OnTwoRoom, it rises monotonically to the deepest layer (layer-5, AUC~$0.925$).
This motivates the task-adaptive heuristic of using layer-0 for Push-T and the deepest predictor layer for TwoRoom (see \Cref{sec:signal_defs}).

\subsection{Pre-Execution Gate and Adaptive Replanning}

\paragraph{Identifying WM dissociation leading to planning failures.}

For each episode, we compute two scalar alignment scores.
The \emph{encoder-centroid cosine} measures cosine similarity between the CEM-predicted encoder centroid and the centroid realized at episode end, quantifying whether the encoder's latent goal representation materialized as planned.
The \emph{predictor-centroid cosine} is the analogous quantity evaluated on predictor centroids, measuring whether the predictor's behavior was oriented toward the planned goal.
A median split on each score partitions episodes into four quadrants: enc$_\text{high}$/pred$_\text{high}$ (both representations aligned); enc$_\text{high}$/pred$_\text{low}$ (goal correctly represented, behavior misaligned); enc$_\text{low}$/pred$_\text{high}$ (behavior superficially oriented toward the goal, but goal representation absent); and enc$_\text{low}$/pred$_\text{low}$ (both fail).
The first quadrant is ``associated''; the latter three are pooled as ``dissociated'' in \Cref{tab:quadrants}.

Associated episodes achieve $45.6\%$ SR on Push-T out-of-distribution and $89.7\%$ on TwoRoom out-of-distribution.
The dissociated episodes pool to $6.4\%$ and $9.9\%$ respectively.

A similar split can be done using activations; however, the centroid split isolates the dissociation more cleanly than activation axes.
A per-episode Pearson correlation between the two split axes is $r=0.185$ for activations vs.\ $r=0.578$ for centroids.

\begin{table}[ht]
\centering
\small
\caption{%
\textbf{Centroid alignment predicts episode outcome in the out-of-distribution regime.}
Episodes are split by median encoder- and predictor-centroid cosine into associated and dissociated groups.
Results are aggregated across five seeds with 50 episodes each.
}\label{tab:quadrants}
\begin{tabular}{lcccc}
\toprule
 & \multicolumn{2}{c}{Push-T} & \multicolumn{2}{c}{TwoRoom} \\
\cmidrule(lr){2-3}\cmidrule(lr){4-5}
 & $n$ & SR & $n$ & SR \\
\midrule
Associated & $79$ & $45.6\%$ & $78$ & $89.7\%$ \\
Dissociated & $171$ & $6.4\%$ & $172$ & $9.9\%$ \\
\bottomrule
\end{tabular}
\end{table}

\paragraph{Behavior-informed adaptive planning.}

Among the dissociated quadrants, enc$_\text{high}$/pred$_\text{low}$ is the only recoverable failure mode, since the encoder correctly represents the goal.
Thus, the failure is purely a predictor of behavioral misalignment.
Meaning, in this instance, a closer goal may lie within the predictor's aligned region.
Crucially, this enc/pred centroid alignment is computable at the first CEM planning call---before any action is taken---making it a deployable pre-execution gate for adaptive replanning.

\begin{table}[ht]
\centering
\small
\caption{%
\textbf{A pre-execution gate identifies dissociated trajectories in a way that can help planning through resampling.}
Gate uses an absolute centroid threshold ($\text{pred}_\text{L0} < 0.997$ for Push-T; $\text{pred}_\text{L1} < 0.993$ for TwoRoom), calibrated from in-distribution data.
Resampling (offset~$50$ instead of~$75$) redirects flagged episodes to a closer goal.
Results are aggregated across five seeds, each with 50 episodes.
}\label{tab:gate_resample}
\begin{tabular}{llrrrr}
\toprule
Task & $n$ flagged & Original Success Rate & Resampled Success Rate & Combined Success Rate \\
\midrule
Push-T  & $72\;(29\%)$ & $\mathbf{0.0\%}$ & $26.4\pm11.1\%$ & $26.4\pm5.0\%$ \\
TwoRoom & $125\;(50\%)$ & $3.8\pm2.1\%$ & $29.6\pm6.7\%$ & $48.0\pm7.3\%$ \\
\bottomrule
\end{tabular}
\end{table}

An absolute threshold on predictor-centroid cosine ($\text{pred}_\text{L0} < 0.997$ for Push-T; $\text{pred}_\text{L1} < 0.993$ for TwoRoom) identifies enc$_\text{high}$/pred$_\text{low}$ episodes pre-execution.
Both thresholds are calibrated from in-distribution data alone with no labeled failures, and yield zero in-distribution false positives on Push-T and a $1\%$ false-alarm rate on TwoRoom.
Flagged episodes are redirected to a closer goal (offset~$50$ instead of~$75$); \Cref{tab:gate_resample} shows the outcome.
On Push-T out-of-distribution, $29\%$ of episodes are flagged, and all fail at the original goal, but $26.4\%$ succeed at the closer goal, lifting the combined success rate by $+7.6$ percentage points.
On TwoRoom out-of-distribution, $50\%$ of episodes are flagged, and resampling increases the success rate by $+13.2$ percentage points.

Crucially, the gains are not a consequence of simply assigning easier goals.
At an average goal offset of $67.8$ (Push-T) and $62.5$ (TwoRoom), a uniform baseline that randomly redirects the same number of episodes to offset~$50$ scores $22.8\%$ and $38.4\%$ combined success rate respectively, versus $26.4\%$ and $48.0\%$ for the gate, confirming that the gate's selectivity adds value beyond uniform easification.

\paragraph{Including centroids in the planning objective.}
Adding centroid alignment terms to the CEM objectives was not found to provide a clear gain.
This is consistent with centroid alignment being a structural property of the WM at the current observation---fixed across the action sequences CEM evaluates---and so it contributes no discriminative signal to guide the action search.
However, incorporating them as a training-time regularization proved fruitful (see \Cref{sec:training}).






\section{Conclusion}

Centroids give a behavioral view of a JEPA WM that complements the knowledge view offered by activations.
We turn this view into a practical, label-free toolkit for runtime monitoring.
With the central focus being a structural failure mode, when the encoder correctly represents the goal but the predictor is not behaviorally oriented toward it, planning almost never succeeds.
This dissociation is identifiable by a simple centroid median split and is isolated far more cleanly by centroid axes than by activation axes.

From this signal follow three capabilities that activations alone do not provide: clean attribution of the encoder--predictor dissociation that predicts failure, a pre-execution gate that recaptures out-of-distribution success through goal resampling ($+7.6$ to $+13.2$\,pp across tasks), and task-salient saliency maps in a single backward pass.
In-distribution, the behavioral and knowledge views are largely interchangeable.
The gap opens in the centroids' favor precisely under distribution shift---where monitoring matters most and where other signals degrade.

All results come from a single WM family (LeWorldModel) on two tasks.
Generalization to other JEPA families, such as DINO-WM~\cite{zhou2025dinowm} and to non-JEPA architectures~\cite{hafner2025dreamerv3,hansen2024tdmpc2}, remains open.

\section*{Acknowledgments}

This work was supported by ONR grant N00014-23-1-2714, DOE grant DE-SC0020345, DOI grant 140D0423C0076, and a Google Cloud Computing Award.

\newpage
\bibliographystyle{unsrtnat}
\bibliography{references}

\newpage
\appendix
\crefalias{section}{appendix}

\section{Signal Definitions}\label{sec:signal_defs}

\paragraph{$c_\text{cen}$ (post-execution calibration).}
After each receding-horizon step, cosine similarity between the predictor's centroid evaluated on the realized next embedding $\bz_t$ and its centroid evaluated on the predicted next embedding $\widehat{\bz}_t$ from the prior step, computed independently per predictor MLP layer.
Low values indicate the predictor failed to anticipate the realized next state.
Layer-0 suffices for Push-T (AUC within $0.02$ of oracle in-distribution, within $0.01$ out-of-distribution), whereas layer-5 (deepest) is consistently oracle-optimal across all seeds and regimes on TwoRoom.
Therefore, a simple task-adaptive heuristic---layer-0 for Push-T, deepest layer for TwoRoom---provides an approximately optimal signal with no post-hoc reasoning.

\paragraph{$c_\text{act}$ (activation calibration).}
Cosine similarity between the encoder embedding $\bz_t$ and the predictor's prior-step embedding prediction $\widehat{\bz}_t$.
The activation-space analog of $c_\text{cen}$.

\paragraph{$c_\text{goal}$ — omp-cen, omp-act, sae-cen, and sae-act (goal-code cosine).}
A feature dictionary of $K\!=\!64$ atoms is built by OMP sparse coding on centroids (omp-cen) or raw activations (omp-act) from 2000 expert trajectories, projected to 64 PCA components.
sae-cen and sae-act use the same preprocessing (StandardScaler and PCA) but replace OMP with a TopK sparse autoencoder ~\cite{bricken2023monosemanticity,templeton2024scaling} ($256$ features and $k\!=\!4$).
sae-cen applies the sparse autoencoder to centroids and sae-act to activations.
The goal-code cosine measures alignment between the current representation and the goal's code.

\paragraph{$c_\text{emb}$ (embedding cost).}
Measures the cosine distance between the predicted final embedding and the goal embedding.

\paragraph{$c_\text{recon}$ (reconstruction error, VAE-style analogue).}
For a given representation type (centroid or activation), $c_\text{recon}$ is the OMP sparse reconstruction error of the current sub-component representation against the training dictionary:
$c_\text{recon} = \|r - D\widehat{\alpha}\|_2$,
where $r$ is the current centroid or activation, $D$ is the $K{=}64$-atom dictionary built from 2000 expert trajectories, and $\widehat{\alpha}$ is the OMP code.
High values indicate the current state is poorly explained by training-distribution features.
This implements the same distributional-anomaly principle as VAE reconstruction-error monitoring~\cite{foundation2025bimanual}---high reconstruction error flags out-of-distribution states---but requires no trained decoder; the same dictionary used for goal-code monitoring (\textit{omp-cen}/\textit{omp-act}) provides the reconstruction signal at no additional cost.
The centroid variant (\textit{recon-cen}) applies $c_\text{recon}$ to the Jacobian centroids of each MLP; the activation variant (\textit{recon-act}) applies it to raw intermediate activations.

\section{Centroid Regularization During Training}\label{sec:training}

We study whether regularizing the centroid geometry during predictor training produces WMs with more structured planning.
Using a frozen DINOv2-small encoder and a 6-layer ViT predictor~\cite{wang2026temporal_straightening}, we compare three curvature regularizers against a reconstruction-only baseline on Point Maze~\cite{fu2020d4rl} and Push-T.
\textsc{cos} applies cosine curvature loss to predictor output embeddings~\cite{wang2026temporal_straightening}; \textsc{centcos} applies the same loss to unit-normalised per-MLP Jacobian centroids; \textsc{cos\_centcos} combines both.

\textsc{cos\_centcos} yields a clear improvement on Point Maze ($+9.0$~pp over baseline, consistent across all 4~seeds), whereas \textsc{centcos} alone shows no reliable benefit ($+0.5$~pp).
On Push-T, all variants fall within $43$--$46\%$, well within noise; binary SR is not discriminating here.
The results suggest output-space and centroid-space regularization are complementary: the benefit appears only when both are combined.

\begin{figure}[ht]
    \centering
    \begin{tikzpicture}
        \begin{groupplot}[
            group style={
                group size=2 by 2,
                horizontal sep=1.8cm, 
                vertical sep=1.2cm,   
            },
            width=0.48\textwidth,
            tick label style={font=\small},
            label style={font=\small},
            title style={font=\small},
        ]
        
        \nextgroupplot[
            height=0.28\textwidth,
            enlargelimits=false,
            y dir=reverse,
            xlabel={Encoder Layer},
            ylabel={Predictor Layer},
            title={Activations},
            point meta min=0.0,
            point meta max=0.42,
            colormap/viridis,
            xmin=0, xmax=11,
            ymin=0, ymax=5
        ]
        \addplot [
            matrix plot,
            mesh/cols=6,
            point meta=explicit
        ] table [
            x=enc_layer, 
            y=pred_layer, 
            meta=act_r2, 
            col sep=comma
        ] {data/exp1_tworooms_heatmaps.csv};

        \nextgroupplot[
            height=0.28\textwidth,
            enlargelimits=false,
            y dir=reverse,
            xlabel={Encoder Layer},
            title={Centroids},
            point meta min=0.0,
            point meta max=0.42,
            colorbar,
            colormap/viridis,
            xmin=0, xmax=11,
            ymin=0, ymax=5
        ]
        \addplot [
            matrix plot,
            mesh/cols=6,
            point meta=explicit
        ] table [
            x=enc_layer, 
            y=pred_layer, 
            meta=cent_r2, 
            col sep=comma
        ] {data/exp1_tworooms_heatmaps.csv};

        \nextgroupplot[
            height=4.5cm,
            xtick={0,1,2,3,4,5},
            xlabel={Layer},
            ylabel={Straightness},
            scaled y ticks=false,
            y tick label style={
                /pgf/number format/.cd,
                fixed,
                precision=3,
                /tikz/.cd,
            }
        ]
        \addplot[color=blue, mark=*, thick] 
            table[x=layer, y=act_straightness_expert, col sep=comma] 
            {data/exp5_tworooms_predictor.csv};

        \addplot[color=blue, mark=*, thick, dashed] 
            table[x=layer, y=act_straightness_shuffled, col sep=comma] 
            {data/exp5_tworooms_predictor.csv};
        
        \addplot[color=red, mark=triangle*, thick] 
            table[x=layer, y=cent_straightness_expert_mean, col sep=comma] 
            {data/exp5_tworooms_predictor.csv};

        \addplot[color=red, mark=triangle*, thick, dashed] 
            table[x=layer, y=cent_straightness_shuffled_mean, col sep=comma] 
            {data/exp5_tworooms_predictor.csv};

        \nextgroupplot[
            height=4.5cm,
            xtick={0,1,2,3,4,5},
            xlabel={Layer},
            ylabel={Autocorrelation},
            scaled y ticks=false,
            y tick label style={
                /pgf/number format/.cd,
                fixed,
                precision=3,
                /tikz/.cd,
            },
            legend style={
                at={(-0.15, -0.25)}, 
                anchor=north,
                legend columns=2,
                font=\small,
            }
        ]
        \addplot[color=blue, mark=*, thick] 
            table[x=layer, y=act_autocorr_expert, col sep=comma] 
            {data/exp5_tworooms_predictor.csv};
        \addlegendentry{Activations}
        
        \addplot[color=red, mark=triangle*, thick] 
            table[x=layer, y=cent_autocorr_expert_mean, y error=cent_autocorr_expert_std, col sep=comma] 
            {data/exp5_tworooms_predictor.csv};
        \addlegendentry{Centroids}

        \addplot[color=blue, mark=*, thick, dashed] 
            table[x=layer, y=act_autocorr_shuffled, col sep=comma] 
            {data/exp5_tworooms_predictor.csv};

        \addplot[color=red, mark=triangle*, thick, dashed] 
            table[x=layer, y=cent_autocorr_shuffled_mean, y error=cent_autocorr_expert_std, col sep=comma] 
            {data/exp5_tworooms_predictor.csv};

        \end{groupplot}
    \end{tikzpicture}
    \caption{
    Reproducing \Cref{fig:r2_heatmaps,fig:trajectory_metrics} for the TwoRooms environment.
    }
    \label{fig:tworooms}
\end{figure}
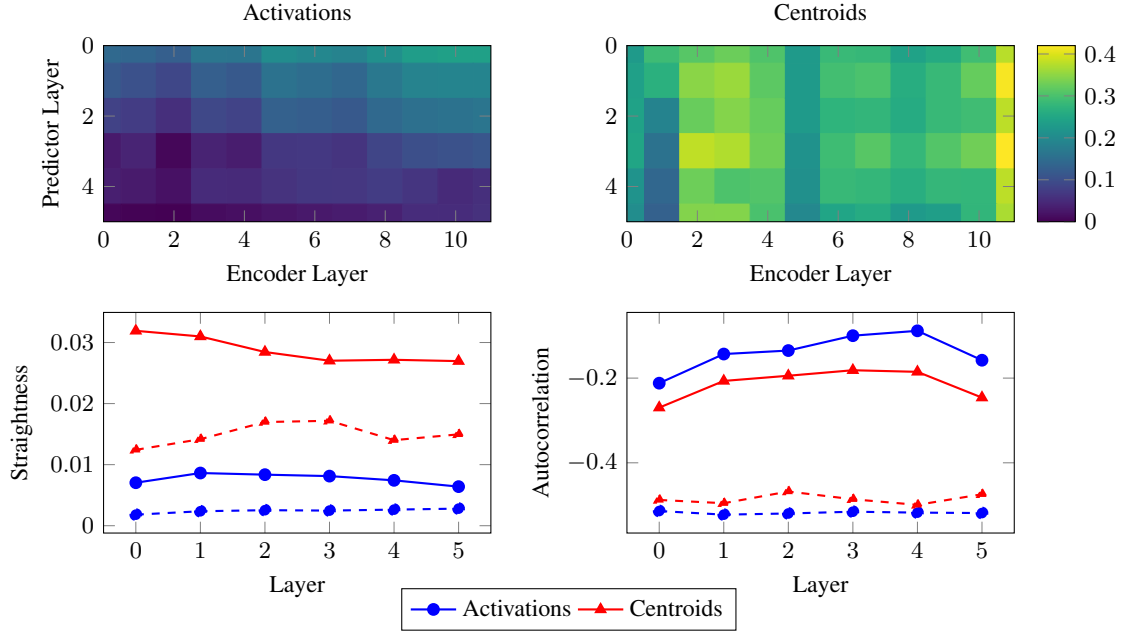

\begin{figure}[ht]
    \centering
    \begin{tikzpicture}
        \begin{groupplot}[
            group style={
                group size=2 by 2,
                horizontal sep=1.8cm, 
                vertical sep=1.2cm,   
            },
            width=0.48\textwidth,
            tick label style={font=\small},
            label style={font=\small},
            title style={font=\small},
        ]
        
        \nextgroupplot[
            height=0.28\textwidth,
            enlargelimits=false,
            y dir=reverse,
            xlabel={Encoder Layer},
            ylabel={Predictor Layer},
            title={Activations},
            point meta min=0.28,
            point meta max=0.95,
            colormap/viridis,
            xmin=0, xmax=11,
            ymin=0, ymax=5
        ]
        \addplot [
            matrix plot,
            mesh/cols=6,
            point meta=explicit
        ] table [
            x=enc_layer, 
            y=pred_layer, 
            meta=act_r2,
            col sep=comma
        ] {data/exp1_reacher_heatmaps.csv};

        \nextgroupplot[
            height=0.28\textwidth,
            enlargelimits=false,
            y dir=reverse,
            xlabel={Encoder Layer},
            title={Centroids},
            point meta min=0.28,
            point meta max=0.95,
            colorbar,
            colormap/viridis,
            xmin=0, xmax=11,
            ymin=0, ymax=5
        ]
        \addplot [
            matrix plot,
            mesh/cols=6,
            point meta=explicit
        ] table [
            x=enc_layer, 
            y=pred_layer, 
            meta=cent_r2,
            col sep=comma
        ] {data/exp1_reacher_heatmaps.csv};
        \nextgroupplot[
            height=4.5cm,
            xtick={0,1,2,3,4,5},
            xlabel={Layer},
            ylabel={Straightness},
            scaled y ticks=false,
            y tick label style={
                /pgf/number format/.cd,
                fixed,
                precision=3,
                /tikz/.cd,
            }
        ]
        \addplot[color=blue, mark=*, thick] 
            table[x=layer, y=act_straightness_expert, col sep=comma] 
            {data/exp5_reacher_predictor.csv};

        \addplot[color=blue, mark=*, thick, dashed] 
            table[x=layer, y=act_straightness_shuffled, col sep=comma] 
            {data/exp5_reacher_predictor.csv};
        
        \addplot[color=red, mark=triangle*, thick] 
            table[x=layer, y=cent_straightness_expert_mean, col sep=comma] 
            {data/exp5_reacher_predictor.csv};

        \addplot[color=red, mark=triangle*, thick, dashed] 
            table[x=layer, y=cent_straightness_shuffled_mean, col sep=comma] 
            {data/exp5_reacher_predictor.csv};

        \nextgroupplot[
            height=4.5cm,
            xtick={0,1,2,3,4,5},
            xlabel={Layer},
            ylabel={Autocorrelation},
            scaled y ticks=false,
            y tick label style={
                /pgf/number format/.cd,
                fixed,
                precision=3,
                /tikz/.cd,
            },
            legend style={
                at={(-0.15, -0.25)}, 
                anchor=north,
                legend columns=2,
                font=\small,
            }
        ]
        \addplot[color=blue, mark=*, thick] 
            table[x=layer, y=act_autocorr_expert, col sep=comma] 
            {data/exp5_reacher_predictor.csv};
        \addlegendentry{Activations}
        
        \addplot[color=red, mark=triangle*, thick] 
            table[x=layer, y=cent_autocorr_expert_mean, y error=cent_autocorr_expert_std, col sep=comma] 
            {data/exp5_reacher_predictor.csv};
        \addlegendentry{Centroids}

        \addplot[color=blue, mark=*, thick, dashed] 
            table[x=layer, y=act_autocorr_shuffled, col sep=comma] 
            {data/exp5_reacher_predictor.csv};

        \addplot[color=red, mark=triangle*, thick, dashed] 
            table[x=layer, y=cent_autocorr_shuffled_mean, y error=cent_autocorr_expert_std, col sep=comma] 
            {data/exp5_reacher_predictor.csv};

        \end{groupplot}
    \end{tikzpicture}
    \caption{
    Reproducing \Cref{fig:r2_heatmaps,fig:trajectory_metrics} for the Reacher environment.
    }
    \label{fig:reacher}
\end{figure}
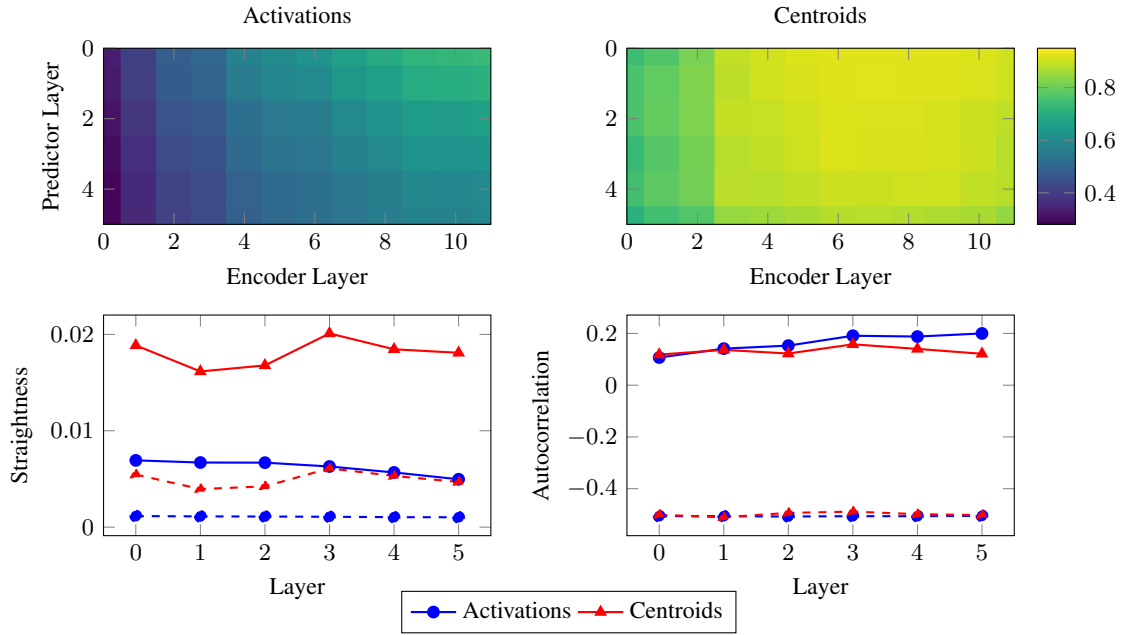

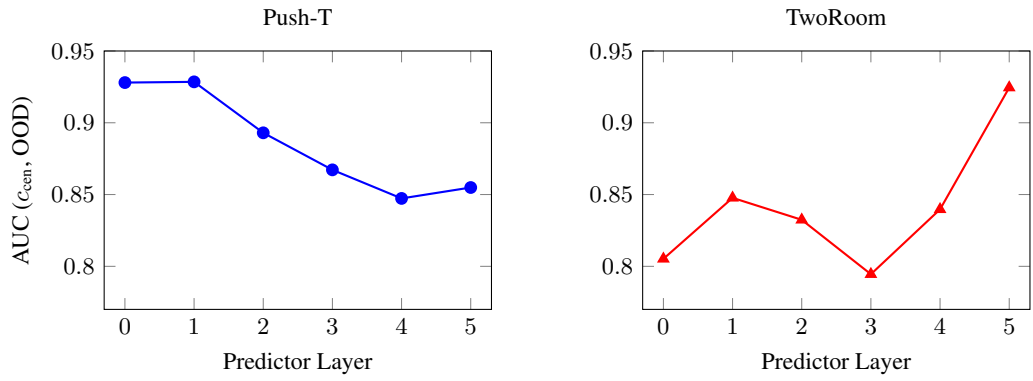
\begin{figure}[ht]
    \centering
    \begin{tikzpicture}
        \begin{groupplot}[
            group style={group size=2 by 1, horizontal sep=2cm},
            width=0.48\textwidth, height=5cm,
            xtick={0,1,2,3,4,5},
            xlabel={Predictor Layer},
            tick label style={font=\small},
            label style={font=\small},
            title style={font=\small},
            xmin=-0.3, xmax=5.3,
            ymin=0.77, ymax=0.95,
        ]
        \nextgroupplot[
            ylabel={AUC ($c_\text{cen}$, OOD)},
            title={Push-T},
        ]
            \addplot[color=blue, mark=*, thick]
                table[x=layer, y=pusht_ood, col sep=comma]{data/sec4_calib_pred_layer_auc.csv};
        \nextgroupplot[
            title={TwoRoom},
        ]
            \addplot[color=red, mark=triangle*, thick]
                table[x=layer, y=tworoom_ood, col sep=comma]{data/sec4_calib_pred_layer_auc.csv};
        \end{groupplot}
    \end{tikzpicture}
    \caption{\textbf{$c_\text{cen}$ AUC by predictor layer in the out-of-distribution regime.}
    Push-T signal peaks at shallow layers (layer-0 and layer-1) and degrades with depth.
    TwoRoom peaks monotonically at the deepest layer (layer-5), motivating the task-adaptive heuristic of using layer-0 for Push-T and the deepest layer for TwoRoom.}
    \label{fig:layer_auc}
\end{figure}

\begin{table}[h]
\centering
\caption{%
\textbf{Planning SR} (4~seeds $\times$ 50~episodes; gradient-descent sub-planner).
Mean $\pm$ std across seeds; per-column best in \textbf{bold}.
}
\label{tab:training_planning}
\small
\begin{tabular}{lcc}
\toprule
Variant & Point Maze SR (\%) & Push-T SR (\%) \\
\midrule
Baseline                         & $82.5 \pm 4.3$          & $\mathbf{46.0} \pm 5.1$ \\
\textsc{cos}                     & $88.0 \pm 3.5$          & $43.0 \pm 3.0$ \\
\textsc{centcos}                 & $83.0 \pm 5.4$          & $\mathbf{46.0} \pm 3.2$ \\
\textsc{cos\_centcos}            & $\mathbf{91.5 \pm 1.7}$ & $44.0 \pm 3.2$ \\
\bottomrule
\end{tabular}
\end{table}

\end{document}